\documentclass[conference]{IEEEtran}
\IEEEoverridecommandlockouts
\usepackage{spconf,amsmath,epsfig}

\let\OLDthebibliography\thebibliography
\renewcommand\thebibliography[1]{
  \OLDthebibliography{#1}
  \setlength{\parskip}{0pt}
  \setlength{\itemsep}{0pt plus 0.3ex}
}
\usepackage{color}
\usepackage{booktabs}
\usepackage{arydshln}
\usepackage{multirow,subfigure}
\usepackage{hyperref,MnSymbol}
\usepackage{cite}
\begin{document}\sloppy

% Example definitions.
% --------------------
\def\x{{\mathbf x}}
\def\L{{\cal L}}

% Title.
% ------
\title{A Perceptual Quality Assessment Exploration for AIGC Images}
%
% Single address.
% ---------------
% \name{ Anonymous submission}
% \address{\\ \\ }
\name{Zicheng Zhang$\tiny{^{1}}$*\thanks{*These authors contributed equally to this work.},Chunyi Li$\tiny{^{1}}$*,Wei Sun$\tiny{^{1}}$,Xiaohong Liu$\tiny{^{1}}$,Xiongkuo Min$\tiny{^{1}}$,and Guangtao Zhai$\tiny{^{1,2}}$}
%Address and e-mail should NOT be added in the submission paper. They should be present only in the camera ready paper. 
\address{$^{1}$Institute of Image Communication and Network Engineering, Shanghai Jiao Tong University\\
 $^{2}$ MoE Key Lab of Artificial Intelligence, AI Institute, Shanghai Jiao Tong University}
% \thanks{This work was supported in part by NSFC (No.62225112, No.61831015), the Fundamental Research Funds for the Central Universities, National Key R\&D Program of China 2021YFE0206700, and Shanghai Municipal Science and Technology Major Project (2021SHZDZX0102).}}
% \name{Zicheng Zhang, Wei Sun, Yingjie Zhou, Wei Lu, Yucheng Zhu, Xiongkuo Min, and Guangtao Zhai}
%Address and e-mail should NOT be added in the submission paper. They should be present only in the camera ready paper. 
% \address{Institute of Image Communication and Network Engineering, Shanghai Jiao Tong University, China}

\maketitle

\begin{abstract}
\underline{AI} \underline{G}enerated \underline{C}ontent (\textbf{AIGC}) has gained widespread attention with the increasing efficiency of deep learning in content creation. AIGC, created with the assistance of artificial intelligence technology, includes various forms of content, among which the AI-generated images (AGIs) have brought significant impact to society and have been applied to various fields such as entertainment, education, social media, etc. However, due to hardware limitations and technical proficiency, the quality of AIGC images (AGIs) varies, necessitating refinement and filtering before practical use. Consequently, there is an urgent need for developing objective models to assess the quality of AGIs. Unfortunately, no research has been carried out to investigate the perceptual quality assessment for AGIs specifically. Therefore, in this paper, we first discuss the major evaluation aspects such as technical issues, AI artifacts, unnaturalness, discrepancy, and aesthetics for AGI quality assessment. Then we present the first perceptual AGI quality assessment database, AGIQA-1K, which consists of 1,080 AGIs generated from diffusion models. A well-organized subjective experiment is followed to collect the quality labels of the AGIs. Finally, we conduct a benchmark experiment to evaluate the performance of current image quality assessment (IQA) models.
\end{abstract}
\section{Introduction}
\label{sec:intro}
\begin{figure}
    \centering
    \includegraphics[width = \linewidth]{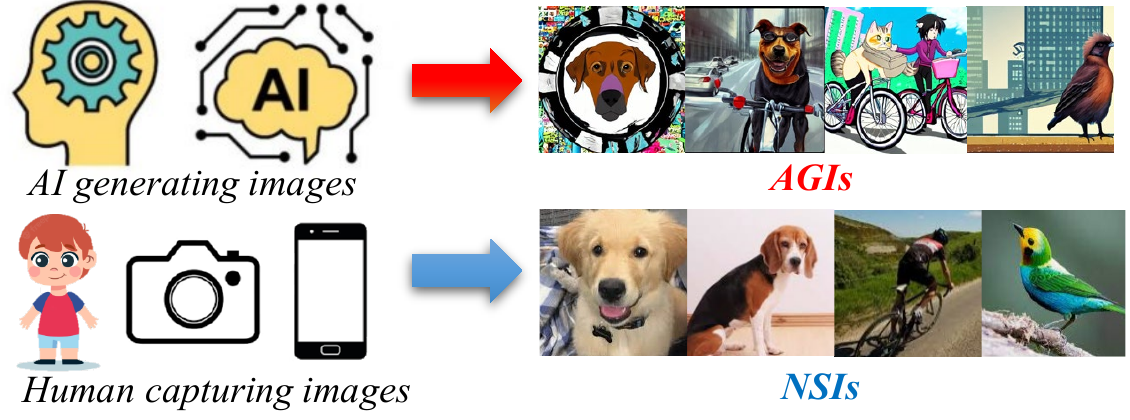}
    \caption{Illustration of the generation process of AGIs and NSIs, where NSIs are captured from the natural scenes and AGIs are directly generated from AI models. }
    \label{fig:spotlight}
\end{figure}

\underline{AI} \underline{G}enerated \underline{C}ontent (\textbf{AIGC}) refers to any form of content, such as text, images, audio, or video, that is created with the help of artificial intelligence technology. With the flourishing development of deep learning, the efficiency of AIGC generation has increased, and AIGC Images (AGIs) are becoming more prevalent in areas such as culture, entertainment, education, social media, etc. Unlike natural scene images (NSIs) that are captured from the natural scenes, AGIs are directly generated from AI models as shown in Fig \ref{fig:spotlight}. Namely, diffusion models \cite{diffusion} and generative adversarial networks \cite{goodfellow2020generative} are capable of generating a great number of images according to our needs. However, due to the hardware limitations and technical proficiency, the quality of AGIs is inconsistent and various, which often requires refinement and filtering before exhibition and being put into practical use. Thus, objective models for evaluating the quality of AGIs are urgently needed. 

During the last decade, large amounts of effort have been put into constructing image quality assessment (IQA) databases and proposing IQA methods for common image contents, such as NSIs \cite{database_NSI}, JPEG2000-compressed \cite{database_Compression}, cartoon \cite{database_Cartoon}, computer-generated \cite{database_CGI}, contrast-changed\cite{database_Contrast}, high dynamic range (HDR) \cite{database_HDR}, in-the-wild \cite{database_wild}, screen content\cite{database_Screen}, and omnidirectional \cite{database_VR} images. All these kinds of images can share some common technical quality assessment dimensions such as illumination, blur, contrast, texture, etc. AGIs obtain some unique quality characteristics and viewers tend to evaluate the quality of AGIs from some different aspects although AGIs are generated under restrictions to be similar to the training images such as NSIs. We summarize some major quality assessment aspects for AGIs here: a) \textbf{\textit{Technical issues}}, which refer to the common distortions that affect the visibility of the image content; b) \textbf{\textit{AI artifacts}}, which indicates the confusing and unexpected components appeared in the images; c) \textbf{\textit{Unnaturalness}}, which stands for the unnaturalness that goes against common sense and the discomfort during the viewing experience; d) \textbf{\textit{Discrepancy}}, which denotes the mismatch extent between the AGIs and our expectation; e) \textbf{\textit{Aesthetics}}, which refers to the overall visual appeal and beauty of the images.

\begin{figure*}[!htp]
    \centering
    \includegraphics[width = \linewidth]{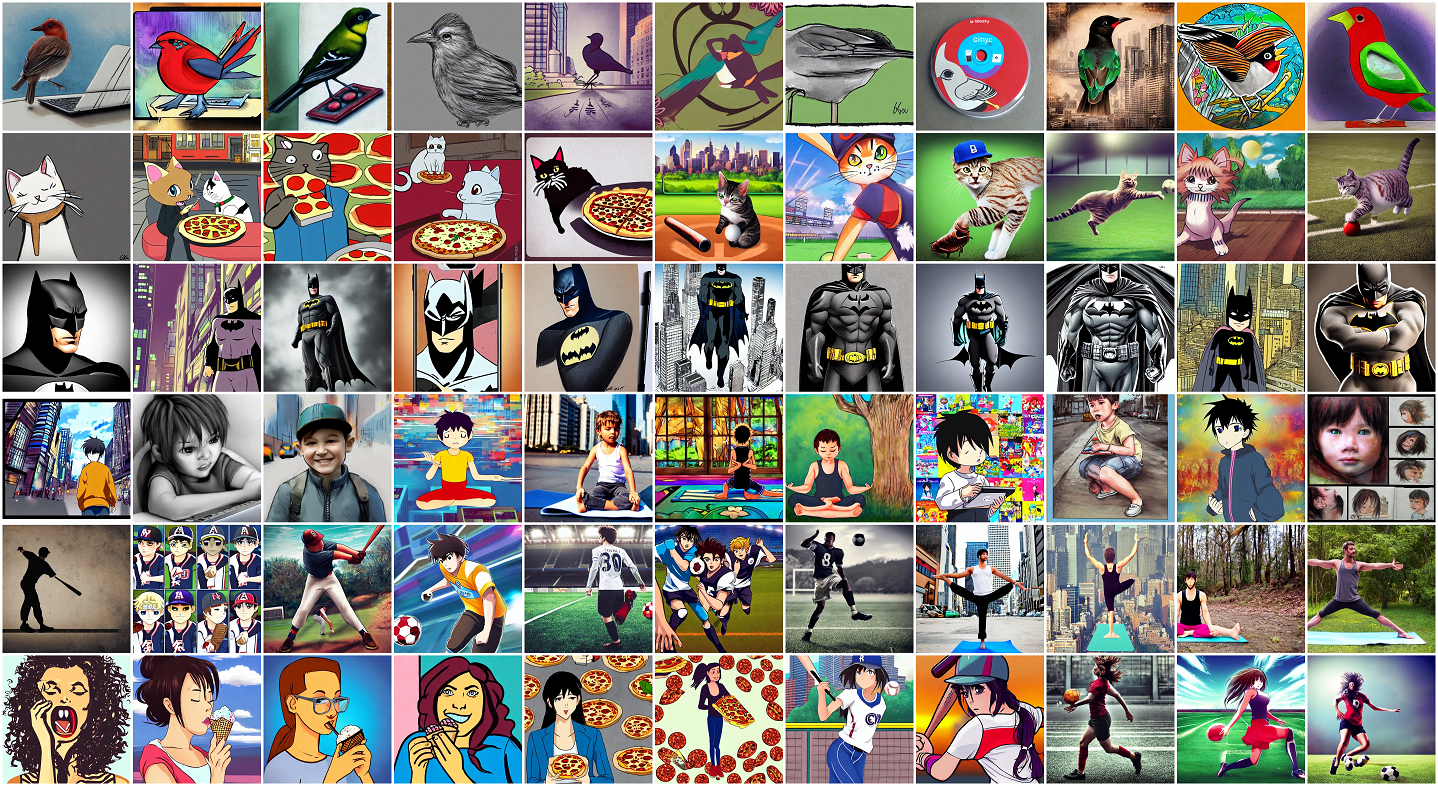}
    \caption{Sample images from the AGIQA-1k database, where the first to sixth rows  show AGIs with (\emph{bird, cat, batman, kid, man, woman}) as the main objects respectively. }
    \label{fig:exhibition}
\end{figure*}

\begin{table}[!tbp]\small
\caption{Illustration of text keywords for generating the AGIs. All the keywords mentioned in the table are used for the \textit{stable-diffusion-v2} while the keywords marked with * are excluded for the \textit{stable-inpainting-v1}.}
\renewcommand\arraystretch{1.2}
\centering
\begin{tabular}{l|c}
\toprule
\textbf{Keywords} & {Content} \\
\hline
\multirow{3}{40pt}{\textbf{Main Objects}} & Bird, Cat, Dog, Batman, \\
&Snoppy, Teddy bear, Kid*, Man, \\
& Woman, Alien, Demon*, Witch* \\
\hline
\multirow{5}{40pt}{\textbf{Second Objects}} & Driving Aircraft*/Bike*/Car*, \\
&Having Hamburger*/Ice-cream*/Pizza,\\
&Playing Baseball/Football/Yoga,\\
&Using CD/Laptop/Phone, \\
&Wearing Coat/Hat/Shirt \\
\hline
\textbf{Places} & City, Wild \\
\hline
\textbf{Styles} & Anime Style, Realistic Style \\
\bottomrule
\end{tabular}
\label{tab:keywords}
\end{table}

However, there has been no scientific research specifically targeted on the perceptual quality of AGIs currently. Therefore, in this paper, we embark on a certain exploration to address the challenge of evaluating the quality of AIGC by carrying out a first-of-a-kind perceptual quality assessment database for AGIs, named AGIQA-1K. Specifically, we employ two latent text-to-image diffusion models \cite{diffusion} \textit{stable-inpainting-v1} and \textit{stable-diffusion-v2} as the AGI models. Then we choose several most popular text keywords from the Internet for AGI generation and a total of 1,080 AGIs are obtained. Afterward, we carry out a subjective experiment in a well-controlled laboratory environment, where the subjects are asked to perceptually evaluate the quality of AGIs following the major quality aspects discussed above. Finally, a benchmark experiment is conducted to evaluate the performance of current IQA models and in-depth discussions are given as well. Our contributions are proposed as follows:
\begin{itemize}
    \item We propose a thorough quality assessment guideline for AGIs, the major evaluation aspects include technical issues, AI artifacts, unnaturalness, discrepancy, and aesthetics.
    \item We are the first to carry out a perceptual AGI quality assessment database (AGIQA-1K), which provides 1,080 AGIs along with quality labels.
    \item A benchmark experiment is conducted to evaluate the performance of current IQA models. 
\end{itemize}

\begin{figure}[!t]
    \centering
    \subfigure[NSI distributions]{\includegraphics[width = 7cm]{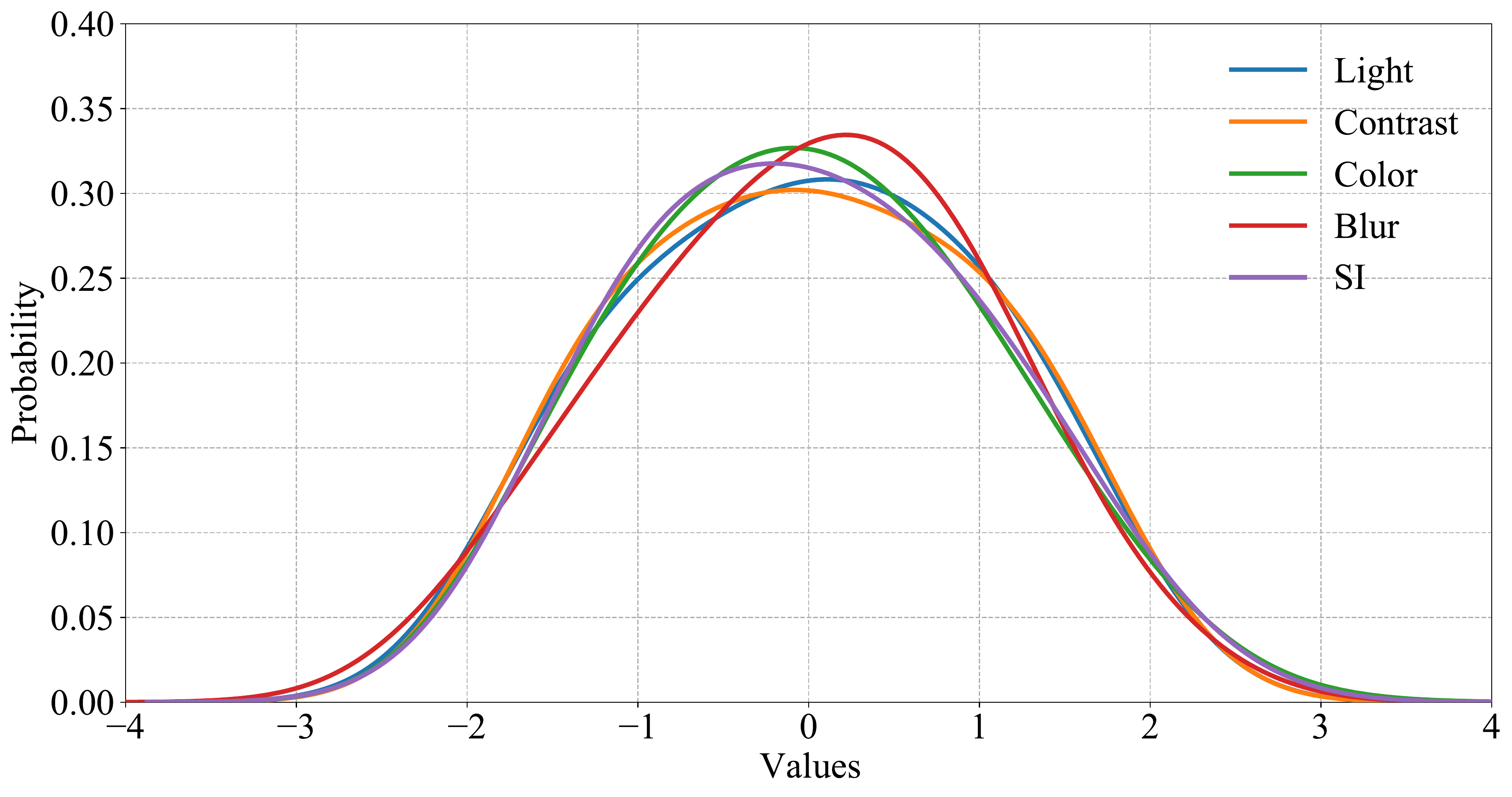}}
    \subfigure[AGI distributions]{\includegraphics[width = 7cm]{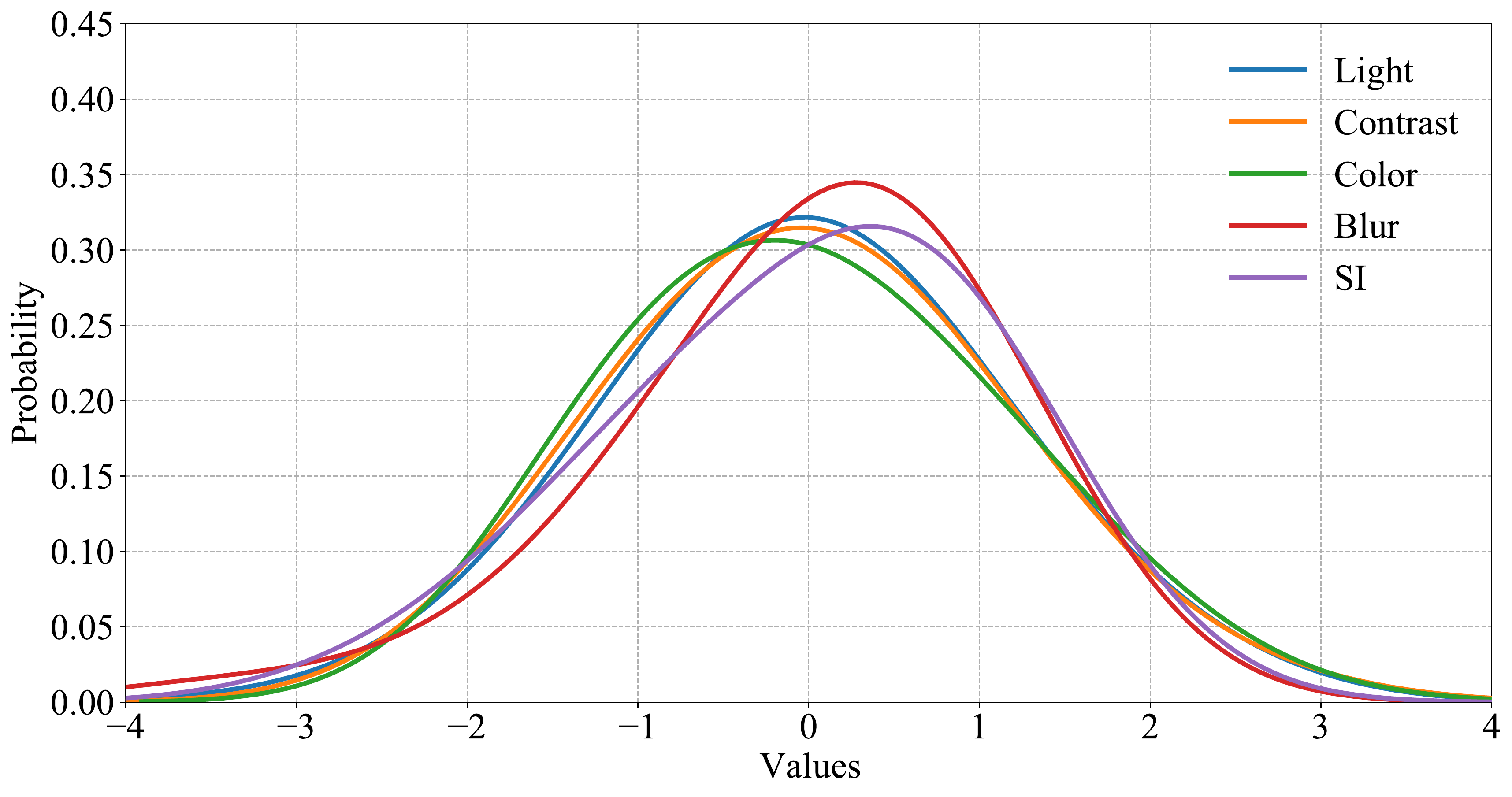}}
    \caption{The normalized probability distributions of the quality-related attributes for NSIs and AGIs. The distributions are obtained from 10,073 NSIs in the KonIQ-10k IQA database \cite{koniq10k} and 1,080 AGIs in the proposed AGIQA-1k database respectively. The 'color' indicates the colorfulness of the images and the 'SI' (spatial information) stands for the content diversity of the images.}
    \label{fig:diff}
\end{figure}

\begin{figure}[!t]
    \centering
    \subfigure[Blur]{\includegraphics[width = 4cm]{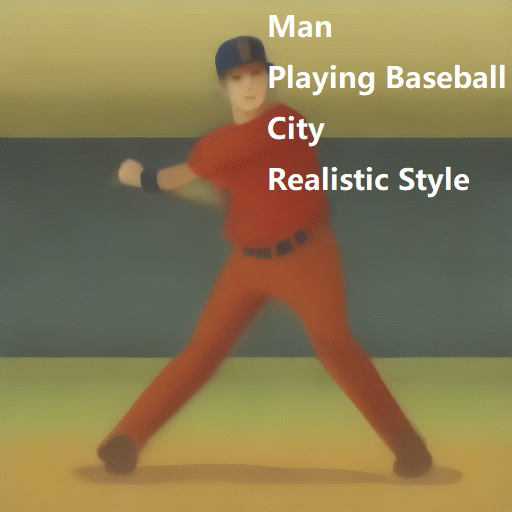}\label{fig:distortions-a}}
    \subfigure[Unexpected artifact]{\includegraphics[width = 4cm]{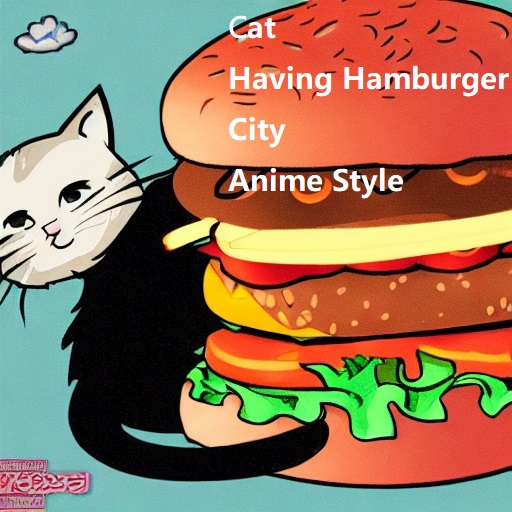}\label{fig:distortions-b}}
    \subfigure[Unnaturalness]{\includegraphics[width = 4cm]{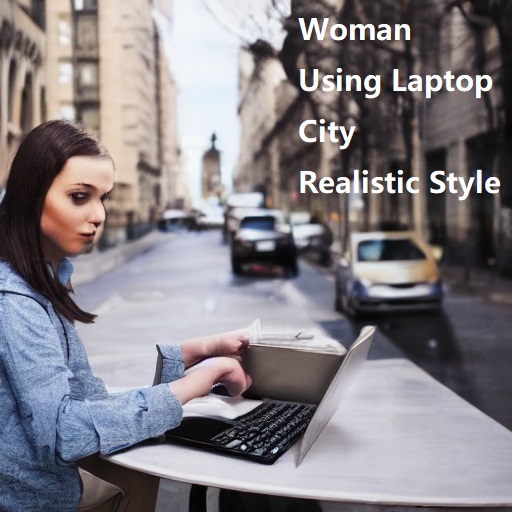}\label{fig:distortions-c}}
    \subfigure[Simple and Text unmatch]{\includegraphics[width = 4cm]{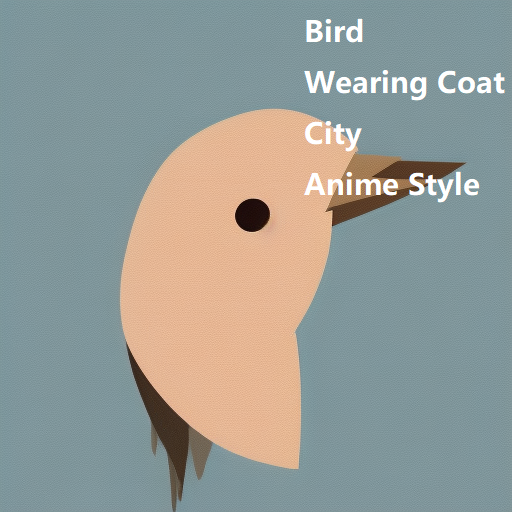}\label{fig:distortions-d}}
    \caption{Exhibition for some common AGI distortions, where the generation keywords are marked in the top right. The content of (a) is poorly visible due to the blur and inexplicit texture. Unexpected artifacts are introduced to the bottom left of (b). (c) contains unnatural content such as the hands of the woman do not cope with common sense. (d) is too simple and does not fit the text keyword of ``Wearing Coat''.}
    \label{fig:distortions}
\end{figure}

\section{Database Construction}

\subsection{AGIs Collection}
Considering the success of stable diffusion models, we select two text-to-image diffusion models \textit{stable-inpainting-v1} and \textit{stable-diffusion-v2} (sub-models derived from \cite{diffusion}) as the AGI models. To ensure content diversity and catch up with the popular trends, we use the hot keywords from the \textit{PNGIMG} website \footnote{https://pngimg.com/} for AGIs generation, and the employed keywords are exhibited in Table \ref{tab:keywords}, which contains the main objects, the second objects, places, and styles. Some sample AGIs are further illustrated in Fig. \ref{fig:exhibition}.

In order to evaluate the statistical discrepancy between NSIs and AGIs, we present the distributions of five quality-related attributes for comparison. The NSIs are sourced from the in-the-wild KonIQ-10k IQA database \cite{koniq10k}, while the AGIs are collected through the proposed AGIQA-1K database. The quality-related attributes under consideration are light, contrast, colorfulness, blur, and spatial information (SI). Detailed descriptions of these attributes can be found in \cite{hosu2017konstanz}. As shown in Fig. \ref{fig:diff}, the quality-related attribute distributions of NSIs and AGIs are quite similar and tend to be Gaussian-like. Specifically, AGIs are relatively blurrier and contain more spatial information than NSIs.

\subsection{Subjective Experiment}
\begin{figure}[!thp]
    \centering
    \includegraphics[width = 0.8\linewidth]{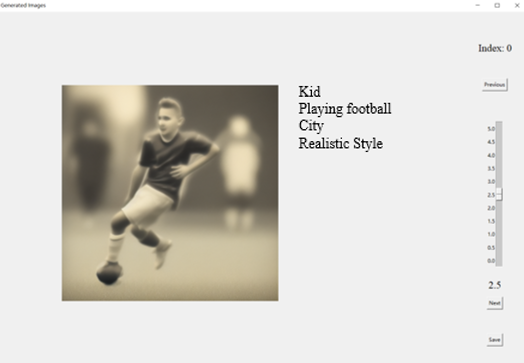}
    \caption{An example of the quality assessment interface, where the AGI and corresponding keywords are shown at the same time. The subject can then evaluate the quality of AGIs and record the quality scores with the scroll bar on the right.}
    \label{fig:interface}
\end{figure}

To evaluate the quality of AGIs, a subjective experiment is conducted following the guidelines of ITU-R BT.500-13 \cite{bt2002methodology}. The subjects are asked to rate the overall quality levels of exhibited AGIs from the technical issues, AI artifacts, unnaturalness, discrepancy, and aesthetic aspects. Some typical distortion examples are shown in Fig. \ref{fig:distortions}. The AGIs are presented in random order on an iMac monitor with a resolution of up to 4096 $\times$ 2304, using an interface designed with Python Tkinter, as shown in Fig. \ref{fig:interface}. The interface allows viewers to browse the previous and next AGIs and rate them using a quality scale that ranges from 0 to 5, with a minimum interval of 0.1. A total of 22 graduate students (10 males and 12 females) participate in the experiment, and they are seated at a distance of around 1.5 times the screen height (45cm) in a laboratory with normal indoor lighting.

To limit the experiment time for each session to less than half an hour, the experiment is split into 5 sessions, each of which includes the subjective quality evaluation for about 200 AGIs.  This results in more than 22$\times$1,080=23,760 quality ratings. 

\subsection{Subjective Data Analysis}
\begin{figure}[!thp]
    \centering
    \includegraphics[width = 0.9\linewidth]{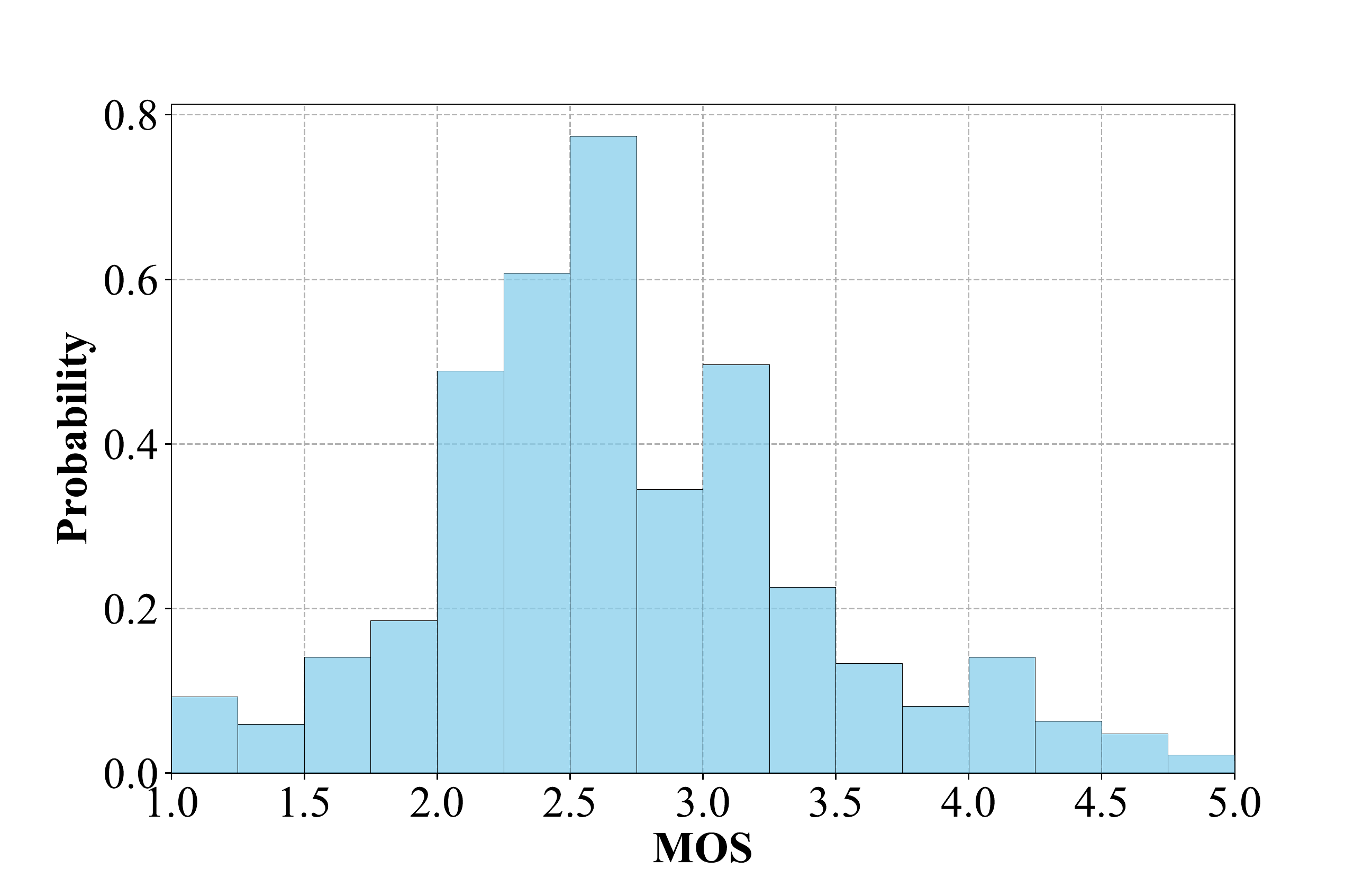}
    \caption{Illustration of the MOS probability distribution.}
    \label{fig:mos}
\end{figure}

After the subjective experiment, all quality ratings from the subjects are collected. The raw rating judged by the $i$-th subject on the $j$-th image is denoted by $r_{ij}$. Z-scores are obtained from the raw ratings using the following formula:
\begin{equation}
z_{ij}=\frac{r_{ij}-\mu_i}{\sigma_i},
\end{equation}
where $\mu_i=\frac{1}{N_i}\sum_{j=1}^{N_i} r_{ij}$, $\sigma_i=\sqrt{\frac{1}{N_i-1}\sum_{j=1}^{N_i}(r_{ij}-\mu_i)^2}$, and $N_i$ is the number of images judged by subject $i$.
Next, ratings from unreliable subjects are removed using the subject rejection procedure recommended by ITU-R BT.500-13 \cite{bt2002methodology}. The mean opinion score (MOS) of image $j$ is computed by averaging the rescaled z-scores:
\begin{equation}
MOS_j=\frac{1}{M}\sum_{i=1}^{M} z_{ij}^{'},
\end{equation}
where $MOS_j$ indicates the MOS for the $j$-th AGI, $M$ is the number of valid subjects, and $z_{ij}^{'}$ are the rescaled z-scores. The corresponding MOS distribution in Fig. \ref{fig:mos} is consistent with previous works \cite{MOS_distribution1} \cite{MOS_distribution2} about subjective diversity.

\section{Experiment}
\subsection{Benchmark Models}
\label{sec:benchmark}
Due to the absence of pristine reference images in the proposed AGIQA-1k database, only no-reference (NR) IQA models are selected for comparison. The selected models can be classified into three groups: 
\begin{itemize}
    \item Handcrafted-based models: This group includes BMPRI \cite{metrics_BMPRI}, CEIQ \cite{metrics_CEIQ}, DSIQA \cite{metrics_DSIQA},  NIQE\cite{metrics_NIQE}, and SISBLIM \cite{metrics_SISBLIM}. These models extract handcrafted features based on prior knowledge about image quality.
    \item Handcrafted \&SVR-based models: This group includes friquee \cite{metrics_friquee}, GMLF \cite{metrics_GMLF}, HIGRADE \cite{metrics_HIGRADE}, NFERM \cite{metrics_NFERM}, and NFSDM \cite{metrics_NFSDM}. These models combine handcrafted features from a Support Vector Regression (SVR) to represent perceptual quality.
    \item Deep learning-based models: This group includes ResNet50 \cite{he2016deep}, StairIQA \cite{metrics_StairIQA}, and MGQA \cite{metrics_MGQA}. These models characterize quality-aware information by training deep neural networks from labeled data.
\end{itemize} 
Notably, the models mentioned above have exhibited strong performance in previous IQA tasks for natural scenes.

%paraphrase the following text in academic style and present tense, and keep the lartex format
\subsection{Evaluation Criteria}
In this study, three primary metrics are utilized to evaluate the consistency between the predicted scores and Mean Opinion Scores (MOSs): Spearman Rank Correlation Coefficient (SRoCC), Pearson Linear Correlation Coefficient (PLCC), Kendall’s Rank Correlation Coefficient (KRoCC). The SRoCC metric measures the similarity between two sets of rankings, while the PLCC metric computes the linear correlation between two groups of rankings. The KRoCC metric, on the other hand, estimates the ordinal relationship between two measured quantities.

To map the predicted scores to MOSs, a five-parameter logistic function is applied, which is a standard practice suggested in \cite{sheikh2006statistical}:
\begin{equation}
\hat{X}=\alpha_{1}\left(0.5-\frac{1}{1+e^{\alpha_{2}\left(X-\alpha_{3}\right)}}\right)+\alpha_{4} X+\alpha_{5},
\end{equation}
where $\{\alpha_{i} \mid i=1,2, \ldots, 5\}$ represent the parameters to be fitted, $y$ and $\hat{y}$ stand for predicted and mapped scores, respectively.

\subsection{Experimental Setup}
All the benchmark models in \ref{sec:benchmark} are validated on the proposed AGIQA-1k database. The database is split randomly in an
80/20 ratio for training/testing while ensuring the image with the same object label falls into the same set. The partitioning and evaluation process is repeated several times for a fair comparison while considering the computational complexity, and the average result is reported as the final performance. For \&SVR-based models, the repeating time is 1,000, implemented by LIBSVM \cite{libsvm} with radial basis function (RBF) kernel. For deep learning-based models, the repeating time is 10, using ResNet50 \cite{he2016deep} as the network backbone. The Adam optimizer \cite{kingma2014adam} (with an initial learning rate of 0.00001 and batch size 40) is used for 100-epochs training on an NVIDIA GTX 4090Ti GPU.

\begin{table*}[!tbhp]\small
\renewcommand\arraystretch{1.1}
\centering
\caption{Performance results on the AGIQA-1k database and two different generative model subsets. The best performance results are marked in {\bf\textcolor{red}{RED}} and the second performance results are marked in {\bf\textcolor{blue}{BLUE}}.}
\label{tab:result}
\begin{tabular}{l|l|ccc|ccc|ccc}
\toprule
\multirow{2}{*}{Metric}                                                                  & Database & \multicolumn{3}{c|}{All}    & \multicolumn{3}{c|}{\textit{stable-inpainting-v1}} & \multicolumn{3}{c}{\textit{stable-diffusion-v2}} \\ \cline{2-11} 
                                                                                         & Corr     & SRoCC   & KRoCC   & PLCC    & SRoCC         & KRoCC        & PLCC         & SRoCC        & KRoCC        & PLCC        \\ \hline
\multirow{5}{*}{\begin{tabular}[c]{@{}l@{}}Hand\\ crafted-\\ based\end{tabular}}         & BMPRI \cite{metrics_BMPRI}    & 0.0651  & 0.0400  & 0.1646  & 0.3746        & 0.2643       & 0.4094       & -0.0158      & -0.0112      & -0.0111     \\
                                                                                         & CEIQ \cite{metrics_CEIQ}     & 0.3069  & 0.2097  & 0.2836  & 0.2348        & 0.1607       & 0.2000       & 0.1314       & 0.0898       & 0.1392      \\
                                                                                         & DSIQA \cite{metrics_DSIQA}    & -0.3047 & -0.2148 & -0.0559 & 0.0428        & 0.0241       & 0.4106       & 0.0046       & 0.0041       & 0.0184      \\
                                                                                         & NIQE \cite{metrics_NIQE}     & -0.5490 & -0.3824 & -0.5048 & 0.0414        & 0.0240       & 0.0712       & -0.2275      & -0.1564      & -0.2392     \\
                                                                                         & SISBLIM \cite{metrics_SISBLIM}  & -0.1309 & -0.0889 & -0.3575 & -0.2410       & -0.1666      & -0.4802      & 0.0541       & 0.0371       & 0.0305      \\ \hline
\multirow{5}{*}{\begin{tabular}[c]{@{}l@{}}Hand\\ crafted\\ \&SVR-\\ based\end{tabular}} & friquee \cite{metrics_NIQE}  & 0.4938  & 0.3469  & 0.4192  & 0.4231        & 0.3024       & 0.3989       & 0.1783       & 0.1244       & 0.2069      \\
                                                                                         & GMLF \cite{metrics_GMLF}     & 0.5575  & 0.4052  & 0.6356  & 0.5062        & 0.3649       & 0.6167       & 0.1501       & 0.1039       & 0.1713      \\
                                                                                         & HIGRADE \cite{metrics_HIGRADE}  & 0.4056  & 0.2860  & 0.4425  & 0.2493        & 0.1732       & 0.2886       & 0.1358       & 0.0943       & 0.1308      \\
                                                                                         & NFERM \cite{metrics_NFERM}    & 0.4540  & 0.3224  & 0.5396  & 0.3874        & 0.2743       & 0.4901       & 0.1193       & 0.0817       & 0.1474      \\
                                                                                         & NFSDM \cite{metrics_NFSDM}    & 0.4314  & 0.3055  & 0.4714  & 0.3840        & 0.2743       & 0.4576       & 0.1002       & 0.0690       & 0.0911      \\ \hline
\multirow{3}{*}{\begin{tabular}[c]{@{}l@{}}Deep\\ learning-\\ based\end{tabular}}        & ResNet50\cite{he2016deep} & {\bf\textcolor{red}{0.6365}}  & {\bf\textcolor{red}{0.4777}}  & {\bf\textcolor{red}{0.7323}}  & {\bf\textcolor{red}{0.6000}}        & {\bf\textcolor{red}{0.4485}}       & {\bf\textcolor{red}{0.7728}}       & {\bf\textcolor{red}{0.3961}}       & {\bf\textcolor{red}{0.2785}}       & {\bf\textcolor{red}{0.4739}}      \\
                                                                                         & StairIQA \cite{metrics_StairIQA} & 0.5504  & 0.4039  & 0.6088  & 0.4669        & 0.2519       & 0.5050       & 0.3486       & 0.2519       & {\bf\textcolor{blue}{0.4186}}      \\
                                                                                         & MGQA \cite{metrics_MGQA}     & {\bf\textcolor{blue}{0.6011}}  & {\bf\textcolor{blue}{0.4456}}  & {\bf\textcolor{blue}{0.6760}}  & {\bf\textcolor{blue}{0.5618}}        & {\bf\textcolor{blue}{0.4250}}       & {\bf\textcolor{blue}{0.7206}}       & {\bf\textcolor{blue}{0.3715}}       & {\bf\textcolor{blue}{0.2584}}       & 0.3593      \\ \bottomrule
\end{tabular}
\end{table*}

\subsection{Performance Discussion}
The performance results on the proposed AGIQA-1K database and corresponding two different generative model subsets are exhibited in Table \ref{tab:result}, from which we can make several conclusions. 1) The handcrafted-based methods achieve poor performance on the whole database and two subsets, which indicates the extracted handcrafted features are not effective for modeling the quality representation of AGIs. This is because most employed handcrafted features of these methods are based on the prior knowledge learned from NSIs, which apparently do not hold for the AGIs. 2) The deep learning-based methods achieve relatively more competitive performance results on the whole database and two subsets. However, they are still far away from satisfactory. 3) Nearly all the IQA models achieve the best performance on the whole database and undergo significant performance drops on the \textit{stable-diffusion-v2} subsets. We attempt to give the reasons for such a phenomenon. More keywords are utilized for the \textit{stable-diffusion-v2} model, therefore making the AGIs generated by a such model more diverse and complicated.  This makes it more challenging for the IQA models to extract quality-aware features from AGIs, which inevitably leads to performance drops.

We further validate the performance of the IQA models on the AGIQA-1K database with the anime and realistic styles. The experimental results are listed in Table \ref{tab:style}. It seems that the IQA models gain similar performance across different styles, which suggests that the styles have a limited impact on the performance of current IQA models. 

\begin{table}[!tbhp]\small
\renewcommand\arraystretch{1.1}
\renewcommand\tabcolsep{4pt}
\centering
\caption{Performance results on the AGIQA-1K database with different styles. The best performance results are marked in {\bf\textcolor{red}{RED}} and the second performance results are marked in {\bf\textcolor{blue}{BLUE}}.}
\vspace{0.1cm}
\label{tab:style}
\begin{tabular}{l|l|cc|cc}
\toprule
\multirow{2}{*}{Metric}                                                                  & Database & \multicolumn{2}{c|}{Anime} & \multicolumn{2}{c}{Realistic} \\ \cline{2-6}
                                                                                         & Corr     & SRoCC         & PLCC         & SRoCC         & PLCC     \\ \cline{1-6}
\multirow{5}{*}{\begin{tabular}[c]{@{}l@{}}Hand\\ crafted-\\ based\end{tabular}}         & BMPRI \cite{metrics_BMPRI}    & 0.1029        & 0.2452       & 0.0257        & 0.1001     \\
                                                                                         & CEIQ \cite{metrics_CEIQ}     & 0.2948        & 0.2746       & 0.3143        & 0.2873      \\
                                                                                         & DSIQA \cite{metrics_DSIQA}    & -0.3282       & -0.0741      & -0.3372       & -0.0857       \\
                                                                                         & NIQE \cite{metrics_NIQE}      & -0.5256       & -0.4901      & -0.5853       & -0.5423      \\
                                                                                         & SISBLIM \cite{metrics_SISBLIM}  & -0.3133       & -0.4331      & -0.0135       & -0.3009     \\ \cline{1-6}
\multirow{5}{*}{\begin{tabular}[c]{@{}l@{}}Hand\\ crafted\\ \&SVR-\\ based\end{tabular}} & friquee\cite{metrics_friquee}   & 0.4654        & 0.3752       & 0.5165        & 0.4153      \\
                                                                                         & GMLF \cite{metrics_GMLF}     & 0.5200        & 0.6338       & {\bf\textcolor{blue}{0.5946}}       & 0.6356      \\
                                                                                         & HIGRADE \cite{metrics_HIGRADE}   & 0.4244        & 0.4895       & 0.4561        & 0.4858        \\
                                                                                         & NFERM \cite{metrics_NFERM}    & 0.4778        & 0.5618       & 0.4834        & 0.5480        \\
                                                                                         & NFSDM \cite{metrics_NFSDM}    & 0.4270        & 0.4768       & 0.3728        & 0.4134          \\ \cline{1-6}
\multirow{3}{*}{\begin{tabular}[c]{@{}l@{}}Deep\\ learning-\\ based\end{tabular}}        & ResNet50 \cite{he2016deep} & 0.5769        & {\bf\textcolor{blue}{0.6769}}       & {\bf\textcolor{red}{0.6686}}        & {\bf\textcolor{red}{0.7577}}        \\
                                                                                         & StairIQA \cite{metrics_StairIQA} & {\bf\textcolor{blue}{0.5947}}        & 0.6385       & 0.5007        & 0.5879        \\
                                                                                         & MGQA \cite{metrics_MGQA}     & {\bf\textcolor{red}{0.6138}}        & {\bf\textcolor{red}{0.6876}}       & 0.5734        & {\bf\textcolor{blue}{0.6613}}        \\ \bottomrule
\end{tabular}
\end{table}

\section{Conclusion}
AIGC has become increasingly popular as deep learning techniques keep improving. However, due to hardware constraints and technical limitations, the quality of AGIs can vary, necessitating refinement and filtering prior to practical usage. Therefore, there is a critical need for developing objective models to assess the quality of AGIs. In this paper, we first discuss significant evaluation aspects, such as technical issues, AI artifacts, unnaturalness, discrepancy, and aesthetics for AGI quality assessment. Then, we carry out the first perceptual AGI quality assessment database, AGIQA-1K, containing 1,080 AGIs generated from diffusion models. A well-organized subjective experiment is conducted to collect quality labels for the AGIs. Subsequently, a benchmark experiment is carried out to evaluate the performance of current IQA models. The experimental results reveal that the current IQA models are not well qualified to deal with AGIQA task and there is still a long way to go.
\bibliographystyle{IEEEbib}
\bibliography{icme2022template}

\begin{thebibliography}{10}

\bibitem{diffusion}
Robin Rombach, Andreas Blattmann, Dominik Lorenz, Patrick Esser, and Björn
  Ommer,
\newblock ``High-resolution image synthesis with latent diffusion models,''
\newblock in {\em IEEE/CVF CVPR}, 2022.

\bibitem{goodfellow2020generative}
Ian Goodfellow, Jean Pouget-Abadie, Mehdi Mirza, Bing Xu, David Warde-Farley,
  Sherjil Ozair, Aaron Courville, and Yoshua Bengio,
\newblock ``Generative adversarial networks,''
\newblock {\em Communications of the ACM}, 2020.

\bibitem{database_NSI}
Zhenyu Peng, Qiuping Jiang, Feng Shao, Wei Gao, and Weisi Lin,
\newblock ``Lggd+: Image retargeting quality assessment by measuring local and
  global geometric distortions,''
\newblock {\em IEEE TCSVT}, 2022.

\bibitem{database_Compression}
Luhong Liang, Shiqi Wang, Jianhua Chen, Siwei Ma, Debin Zhao, and Wen Gao,
\newblock ``No-reference perceptual image quality metric using gradient
  profiles for jpeg2000,''
\newblock {\em Signal Processing: Image Commun}, 2010.

\bibitem{database_Cartoon}
Chunyi Li, Zicheng Zhang, Wei Sun, Xiongkuo Min, and Guangtao Zhai,
\newblock ``A full-reference quality assessment metric for cartoon images,''
\newblock in {\em IEEE MMSP}, 2022.

\bibitem{database_CGI}
Zicheng Zhang, Wei Sun, Xiongkuo Min, Tao Wang, Wei Lu, and Guangtao Zhai,
\newblock ``Distinguishing computer-generated images from photographic images:
  a texture-aware deep learning-based method,''
\newblock in {\em IEEE VCIP}, 2022.

\bibitem{database_Contrast}
Shiqi Wang, Kede Ma, Hojatollah Yeganeh, Zhou Wang, and Weisi Lin,
\newblock ``A patch-structure representation method for quality assessment of
  contrast changed images,''
\newblock {\em IEEE SPL}, 2015.

\bibitem{database_HDR}
Ke~Gu, Shiqi Wang, Guangtao Zhai, Siwei Ma, Xiaokang Yang, Weisi Lin, Wenjun
  Zhang, and Wen Gao,
\newblock ``Blind quality assessment of tone-mapped images via analysis of
  information, naturalness, and structure,''
\newblock {\em IEEE TMM}, 2016.

\bibitem{database_wild}
Zhenqiang Ying, Haoran Niu, Praful Gupta, Dhruv Mahajan, Deepti Ghadiyaram, and
  Alan Bovik,
\newblock ``From patches to pictures (paq-2-piq): Mapping the perceptual space
  of picture quality,''
\newblock in {\em IEEE/CVF CVPR}, 2020.

\bibitem{database_Screen}
Shiqi Wang, Ke~Gu, Xinfeng Zhang, Weisi Lin, Siwei Ma, and Wen Gao,
\newblock ``Reduced-reference quality assessment of screen content images,''
\newblock {\em IEEE TCSVT}, 2018.

\bibitem{database_VR}
Mai Xu, Chen Li, Shanyi Zhang, and Patrick~Le Callet,
\newblock ``State-of-the-art in 360° video/image processing: Perception,
  assessment and compression,''
\newblock {\em IEEE JSTSP}, 2020.

\bibitem{koniq10k}
V.~{Hosu}, H.~{Lin}, T.~{Sziranyi}, and D.~{Saupe},
\newblock ``Koniq-10k: An ecologically valid database for deep learning of
  blind image quality assessment,''
\newblock {\em IEEE TIP}, 2020.

\bibitem{hosu2017konstanz}
Vlad Hosu, Franz Hahn, Mohsen Jenadeleh, Hanhe Lin, Hui Men, Tam{\'a}s
  Szir{\'a}nyi, Shujun Li, and Dietmar Saupe,
\newblock ``The konstanz natural video database (konvid-1k),''
\newblock in {\em IEEE QoMEX}, 2017.

\bibitem{bt2002methodology}
I.~T. Union,
\newblock ``Methodology for the subjective assessment of the quality of
  television pictures,''
\newblock {\em ITU-R Recommendation BT. 500-11}, 2002.

\bibitem{MOS_distribution1}
Yixuan Gao, Xiongkuo Min, Yucheng Zhu, Jing Li, Xiao-Ping Zhang, and Guangtao
  Zhai,
\newblock ``Image quality assessment: From mean opinion score to opinion score
  distribution,''
\newblock in {\em ACM MM}, 2022.

\bibitem{MOS_distribution2}
Hossein Talebi and Peyman Milanfar,
\newblock ``Nima: Neural image assessment,''
\newblock {\em IEEE TIP}, 2018.

\bibitem{metrics_BMPRI}
Xiongkuo Min, Guangtao Zhai, Ke~Gu, Yutao Liu, and Xiaokang Yang,
\newblock ``Blind image quality estimation via distortion aggravation,''
\newblock {\em IEEE TBC}, 2018.

\bibitem{metrics_CEIQ}
Jia Yan, Jie Li, and Xin Fu,
\newblock ``No-reference quality assessment of contrast-distorted images using
  contrast enhancement,''
\newblock {\em arXiv preprint arXiv:1904.08879}, 2019.

\bibitem{metrics_DSIQA}
Niranjan~D Narvekar and Lina~J Karam,
\newblock ``A no-reference perceptual image sharpness metric based on a
  cumulative probability of blur detection,''
\newblock in {\em QoMEX}, 2009.

\bibitem{metrics_NIQE}
Anish Mittal, Rajiv Soundararajan, and Alan~C Bovik,
\newblock ``Making a “completely blind” image quality analyzer,''
\newblock {\em IEEE SPL}, 2012.

\bibitem{metrics_SISBLIM}
Ke~Gu, Guangtao Zhai, Xiaokang Yang, and Wenjun Zhang,
\newblock ``Hybrid no-reference quality metric for singly and multiply
  distorted images,''
\newblock {\em IEEE TBC}, 2014.

\bibitem{metrics_friquee}
Deepti Ghadiyaram and Alan~C Bovik,
\newblock ``Perceptual quality prediction on authentically distorted images
  using a bag of features approach,''
\newblock {\em Journal of Vision}, 2017.

\bibitem{metrics_GMLF}
Wufeng Xue, Xuanqin Mou, Lei Zhang, Alan~C Bovik, and Xiangchu Feng,
\newblock ``Blind image quality assessment using joint statistics of gradient
  magnitude and laplacian features,''
\newblock {\em IEEE TIP}, 2014.

\bibitem{metrics_HIGRADE}
D~Kundu, D~Ghadiyaram, AC~Bovik, and BL~Evans,
\newblock ``Large-scale crowdsourced study for high dynamic range images,''
\newblock {\em IEEE TIP}, 2017.

\bibitem{metrics_NFERM}
Ke~Gu, Guangtao Zhai, Xiaokang Yang, and Wenjun Zhang,
\newblock ``Using free energy principle for blind image quality assessment,''
\newblock {\em IEEE TMM}, 2014.

\bibitem{metrics_NFSDM}
Ke~Gu, Guangtao Zhai, Xiaokang Yang, Wenjun Zhang, and Longfei Liang,
\newblock ``No-reference image quality assessment metric by combining free
  energy theory and structural degradation model,''
\newblock in {\em IEEE ICME}, 2013.

\bibitem{he2016deep}
Kaiming He, Xiangyu Zhang, Shaoqing Ren, and Jian Sun,
\newblock ``Deep residual learning for image recognition,''
\newblock in {\em IEEE/CVF CVPR}, 2016.

\bibitem{metrics_StairIQA}
Wei Sun, Huiyu Duan, Xiongkuo Min, Li~Chen, and Guangtao Zhai,
\newblock ``Blind quality assessment for in-the-wild images via hierarchical
  feature fusion strategy,''
\newblock in {\em IEEE BMSB}, 2022.

\bibitem{metrics_MGQA}
Tao Wang, Wei Sun, Xiongkuo Min, Wei Lu, Zicheng Zhang, and Guangtao Zhai,
\newblock ``A multi-dimensional aesthetic quality assessment model for mobile
  game images,''
\newblock in {\em IEEE VCIP}, 2021.

\bibitem{sheikh2006statistical}
Hamid~R Sheikh, Muhammad~F Sabir, and Alan~C Bovik,
\newblock ``A statistical evaluation of recent full reference image quality
  assessment algorithms,''
\newblock {\em IEEE TIP}, 2006.

\bibitem{libsvm}
Chih-Chung Chang and Chih-Jen Lin,
\newblock ``Libsvm: a library for support vector machines,''
\newblock {\em ACM TIST}, 2011.

\bibitem{kingma2014adam}
Diederik~P Kingma and Jimmy Ba,
\newblock ``Adam: A method for stochastic optimization,''
\newblock in {\em ICLR}, 2015.

\end{thebibliography}

\end{document}